\documentclass[runningheads]{llncs}
\usepackage[T1]{fontenc}
\usepackage{graphicx}
\usepackage[table,xcdraw]{xcolor}
\usepackage{multirow}
\usepackage{amsmath}
\usepackage{amssymb}
\usepackage{bm}
\usepackage{upgreek}
\usepackage{adjustbox}
\usepackage{makecell}
\usepackage{arydshln}

\usepackage{amsmath,amssymb,amsfonts}
\usepackage{graphicx}
\usepackage{textcomp}
\usepackage{xcolor}
\usepackage{multirow}
\usepackage{adjustbox}
\usepackage{makecell}
\usepackage{algpseudocode}

\usepackage{wrapfig}
\usepackage{marvosym}
\newcommand\blfootnote[1]{%
  \begingroup
  \renewcommand\thefootnote{}\footnote{#1}%
  \addtocounter{footnote}{-1}%
  \endgroup
}

\colorlet{blue}{black}

\begin{document}
\title{Structure Observation Driven Image-Text Contrastive Learning for Computed Tomography Report Generation}
\titlerunning{Structural-wise image-text contrastive learning for CT report generation}
%
\author{Hong Liu\inst{2,3} 
\and Dong Wei\inst{3} 
\and Qiong Peng\inst{1} 
\and Yawen Huang\inst{3} 
\and Xian Wu\inst{3}\textsuperscript{(\Letter)} 
\and\\ Yefeng Zheng\inst{3,4}\textsuperscript{(\Letter)} 
\and Liansheng Wang\inst{1,2}\textsuperscript{(\Letter)}
} 
%
\authorrunning{H. Liu et al.}
%
\institute{School of Informatics, Xiamen University, Xiamen, China\\
\and
National Institute for Data Science in Health and Medicine, Xiamen University\\ \email{lswang@xmu.edu.cn, \{liuhong,qpeng\}@stu.xmu.edu.cn}
\and
Jarvis Research Center, Tencent YouTu Lab, Shenzhen, China\\
\email{\{donwei,yawenhuang,kevinxwu,yefengzheng\}@tencent.com}
\and
Medical Artificial Intelligence Laboratory, Westlake University, Hangzhou, China\\
\email{zhengyefeng@westlake.edu.cn}
}
%
\maketitle              
\begin{abstract}
Computed Tomography Report Generation (CTRG) aims to automate the clinical radiology reporting process, thereby reducing the workload of report writing and facilitating patient care.\blfootnote{H. Liu and D. Wei---—Contributed equally; H. Liu contributed to this work during an internship at Tencent.}
While deep learning approaches have achieved remarkable advances in X-ray report generation, their effectiveness may be limited in CTRG due to larger data volumes of CT images and more intricate details required to describe them.
This work introduces a novel two-stage (structure- and report-learning) framework tailored for CTRG featuring effective structure-wise image-text contrasting.
In the first stage, a set of learnable \textit{structure-specific visual queries} ``observe'' corresponding structures in a CT image.
The resulting observation tokens are contrasted with structure-specific textual features extracted from the accompanying radiology report with a structure-wise image-text contrastive loss. 
In addition,
text-text similarity-based soft pseudo targets are proposed to mitigate the impact of false negatives, i.e., semantically identical image structures and texts from non-paired images and reports.
Thus, the model learns structure-level semantic correspondences between CT images and reports.
Further, a dynamic, diversity-enhanced negative queue is proposed to guide the network in learning to discriminate various abnormalities.
In the second stage, the visual structure queries are frozen and used to select the critical image patch embeddings depicting each anatomical structure, minimizing distractions from irrelevant areas while reducing memory consumption.
Also, a text decoder is added and trained for report generation.
Our extensive experiments on two public datasets demonstrate that our framework establishes new state-of-the-art performance for CTRG in clinical efficiency, 
and its components are effective. 

\keywords{Computed tomography report generation  \and Structure-wise image-text contrastive learning \and Text-text similarity soft pseudo targets.}
\end{abstract}
%
%
\begin{figure}[tp]
    \centering
    \includegraphics[width=0.87\linewidth,trim=12 0 13 0,clip]{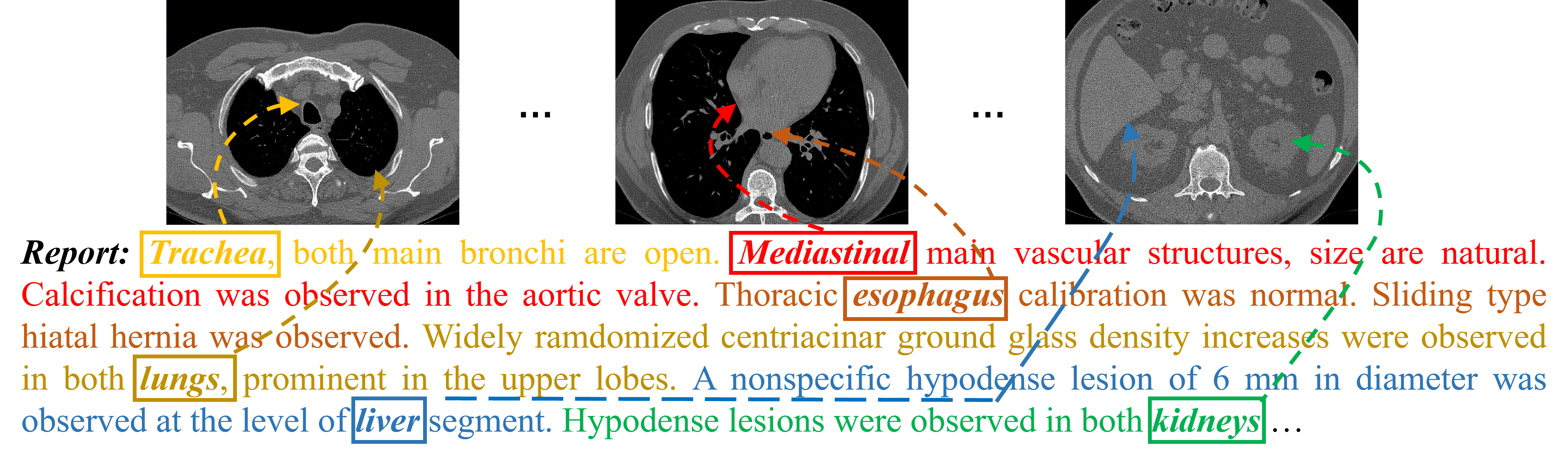}\vspace{-1.25mm}
    \caption{An example CT image and corresponding report demonstrating highly structured text descriptions and visual correspondences.
}\label{data_vis}
\end{figure}

\section{Introduction}
Diagnostic reports of widely used radiographs, such as X-ray and computed tomography (CT), play a crucial role in clinical practice, serving as essential documentation for healthcare and facilitating accurate diagnosis and timely treatment. 
Meanwhile, medical image interpretation requires domain expertise and careful observation, facing the potential risk of misdiagnosis in the presence of staff shortage and excessive workload \cite{brady2012discrepancy}.

Many approaches have attempted automatic medical report generation for 2D X-ray images with deep learning (DL) techniques \cite{chen2020generating,jing2020show,li2019knowledge,li2018hybrid}.
These approaches have achieved promising advancements, but their transferability to 3D radiographs, such as CT, has yet to be validated.
%
The task of CT report generation (CTRG) is more complex than X-ray report generation, mainly because of the combination of two factors:
(1) substantially more data to process: e.g., a typical chest X-ray image is 512$\times$512 to 1024$\times$1024 pixels, whereas a typical thin-slice chest CT volume may contain a few hundred slices of 512$\times$512 pixels; and 
(2) substantially more information to interpret: e.g., chest X-rays typically reveal tens of main findings \cite{irvin2019chexpert}, whereas chest CT images may concern 
more than 80 abnormalities \cite{draelos2021machine}.
%
%
As of now, CTRG is drawing increasing research interest.
Tang {et al.} \cite{tang2024work} proposed a self-attention-based scan localizer framework (SL-DG) for CTRG.
To tackle the above-described challenges, SL-DG trained a scan localizer network based on disease classification using manually labeled medical terms as supervision, to select the critical slices in a CT sequence for report generation.
However, manually labeling medical terms for CT images was labor-intensive.
In addition, the limited supervision restricted the model's generality and usability, as only a few major abnormalities were annotated.
Further, the 2D slice-based approach overlooked the inherent 3D context of CT volumes.

%

This work proposes a two-stage framework for CTRG featuring a novel structure-learning stage, where the model is trained with our proposed structure-level abnormality-enhanced contrastive learning objective.
This objective is driven by anatomical structure observations derived from radiology reports, such that the model learns good representations for main structures in CT images.
Unlike existing methods relying on well-prepared prior knowledge such as knowledge graphs \cite{chen2020generating,liu2021exploring} or manual abnormality annotations \cite{tang2024work}, our framework only requires high-level generic knowledge about \textit{what} anatomical structures that a CT image depicts, e.g., a chest CT mainly depicts thorax, rib, lung, heart, pleura, liver, kidney, thyroid, etc.
Fig. \ref{data_vis} shows a CT image and corresponding report, demonstrating highly structured text descriptions and visual correspondences.
Based on this knowledge, we propose to extract the most informative representations for each structure with a set of learnable structure-specific \textit{visual queries} via cross-attention.
Thus, the training becomes more focused and less computationally demanding.
To leverage a broad source of supervision from the reports, we use a pretrained text encoder to extract structure-specific features from corresponding sentences. Then, learning of the structure-specific queries is supervised by a structure-wise image-text contrastive loss.
In addition, we propose text-text similarity-based soft pseudo targets to mitigate the impact of false negatives, i.e., semantically identical image structures and texts from non-paired images and reports.
%
A diversity-enhanced negative queue update strategy is proposed to guide the network in learning to discriminate various abnormalities.
%
%
%
In the second stage, the learned visual encoder and structural queries are frozen.
A text decoder is added and trained for report generation.
The frozen visual structural queries extract structural representations and select the most representative image patch embeddings as the decoder's input.
Extensive experiments on two public CTRG datasets demonstrate our framework's superior performance to existing state-of-the-art methods and the efficacy of its novel designs.
\vspace{-2.5mm}
\section{Related Work}

\vspace{-1mm}
\subsubsection{Radiology Report Generation.}


Early studies directly applied various encoder-decoder models such as CNN-RNN \cite{vinyals2015show}, LRCN \cite{donahue2015long}, and AdaAtt \cite{lu2017knowing}.
However, the lack of domain knowledge in these models limited their performance.
%
To address this issue, Chen et al. \cite{chen2020generating} and Liu et al. \cite{liu2021exploring} proposed incorporating medical prior knowledge into network training with a memory-driven transformer and a medical knowledge graph, respectively.
%
Similarly, Li et al. \cite{li2023auxiliary} extracted normal and abnormal terminologies from the reports and used them as nodes to build a knowledge graph driving the report generation.
In contrast, Yang et al. \cite{yang2022knowledge} introduced general and input-specific knowledge into training with a novel knowledge-enhanced multi-head attention mechanism.
Although these methods made notable progress, the requirement of well-prepared/processed prior knowledge (e.g., in the form of a knowledge graph) limited their extensibility.

Another research trend is to improve the extraction of visual representations.
Region-wise image features \cite{ren2015faster} have shown their effectiveness for fine-grained tasks, including report generation \cite{tanida2023interactive}.
Yet training region-detection networks required a considerable annotation burden and complicated the process.
Wang et al. \cite{wang2021self} introduced an image-text matching branch to extract text-correlated visual features.
Meanwhile, self-supervised pretraining methods have demonstrated strong representation learning capability \cite{chan2020cocon,gong2022self} and efficacy in medical report generation \cite{10372071}.
The methods above have made remarkable progress in generating medical reports for 2D radiographs, typically chest X-rays. However, the transferability to 3D radiographs, such as CT, has yet to be validated.

Recently, researchers explored integrating semantic information to improve report generation through multi-task learning.
Wang et al. \cite{wang2022medical} trained a multi-label classification network to guide report generation by extracting 768 high-frequency medical terms from RadGraph \cite{jain2021radgraph}. 
Jin et al. \cite{jin2024promptmrg} proposed explicitly guiding the generation process with token prompts derived from diagnostic results obtained through classification.
For 3D CTRG, Hamamci et al. \cite{hamamci2024ct2rep} proposed enhancing the model with a cross-attention multimodal fusion module and hierarchical memory to incorporate longitudinal multimodal data.
Chen et al. \cite{chen2024dia} introduced a disease prototype memory bank and a disease-aware attention module to capture diagnostic information.
This information was used as guidance prompts to the LLaMA2-7B \cite{touvron2023llama} model. 
Tang et al. \cite{tang2024work} proposed a self-attention-based scan localizer framework (SL-DG) for CTRG.
SL-DG involved training a scan localizer network with disease classification. 
The network was supervised by manually labeled medical terms, and was used to select critical slices in a CT sequence for report generation. 
While incorporating diagnostic results significantly improved the accuracy of generated reports, the limited supervision restricted the model's generality and usability, as only a few major abnormalities were annotated.
In contrast, this work proposes directly extracting structure-level diagnostic information from raw reports via structure observation-driven image-text contrastive learning.
Leveraging detailed structural information for contrastive learning, it aims to enhance the ability to generate accurate and comprehensive reports and improve the generality and usability of the model.

\begin{figure*}[t]
    \centering
    \includegraphics[trim=0 0 0 0, clip, width=.90\linewidth]{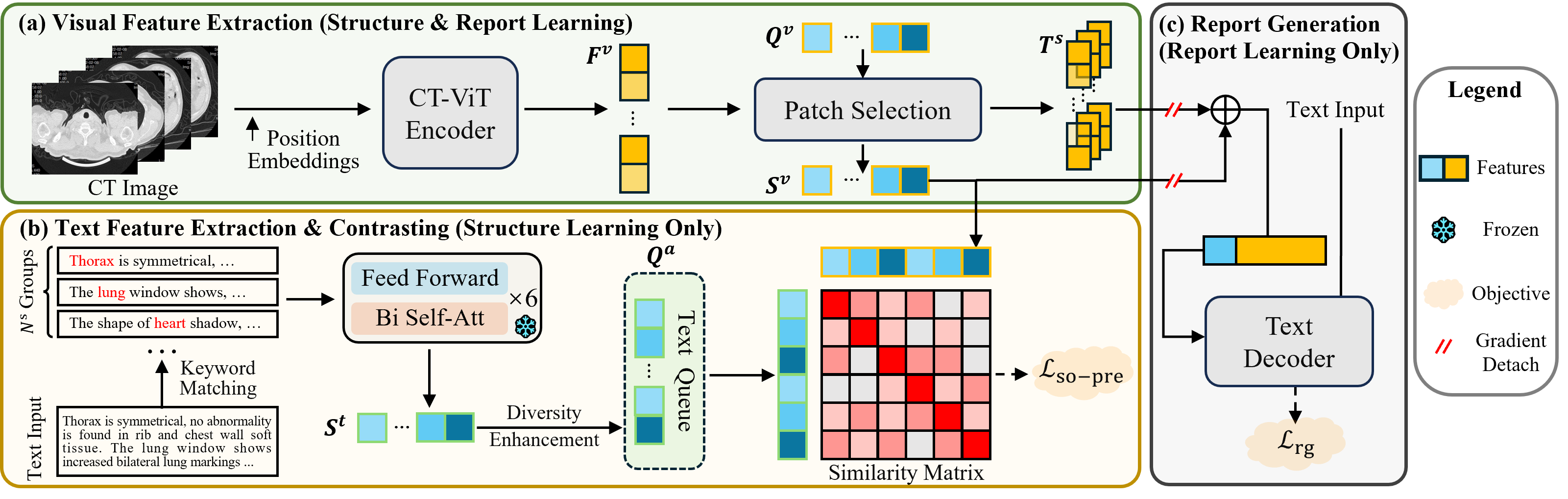}
    \vspace{-1mm}
    \caption{Overview of our proposed framework.}
    \label{framework}\vspace{-2.75mm}
\end{figure*}

\vspace{3.5mm}\noindent\textbf{Medical Image-Language Contrastive Learning.}
Contrastive learning (CL) \cite{chen2020simple,he2020momentum} aims to train models to distinguish between similar and dissimilar sample pairs by minimizing the distances between positive pairs and maximizing the distances between negative ones in the representation space. 
To improve the efficiency of CL, some works proposed to exploit the cross-modal local level \cite{huang2021gloria} and disease-prototype level \cite{wang2022multi} semantic consistency. 
To mitigate the impact of false negatives, Byun et al. \cite{byun2024mafa} proposed MAFA (MAnaging FAlse negatives) with connection mining to convert false negatives into positives and label smoothing for image-text CL loss.
In the medical image domain, CL has been well explored and achieved encouraging results \cite{chaitanya2020contrastive,feng2020parts2whole,liu2023improving,taleb2022contig}.
%
Liu et al. \cite{liu2023improving} proposed triaging image-report pairs into positive, negative, and additional neutral groups based on \textit{global} inter-report similarity, to reduce false-negative
pairs (i.e., semantically similar but from different cases) in contrastive pretraining.
%
In contrast, we propose to learn \textit{structure-wise local} image representations from reports via cross-modal structure-level image-text contrastive learning.
\section{Method}

\subsubsection{Overview.} 
Fig. \ref{framework} shows an overview of our framework, including a structure- and a report-learning stage.
%
As the premise, we use the CT-ViT \cite{hamamci2023generatect} to extract image patch embeddings.
In the structure-learning stage, a set of $N^s$ learnable structure-specific visual queries $Q^v$ extracts a set of subject-specific structure observation tokens $S^v$ with an image patch selection layer.
Meanwhile, $N^s$ groups of structure-specific sentences are fed into a pretrained BERT-based model \cite{hamamci2024foundation} to extract corresponding textual structure observation tokens $S^t$.
%
%
To enforce consistency between the visual and textual structures, 
a structure observation driven image-text contrasting loss $\mathcal{L}_\text{so-pre}$ is enforced between $S^v$ and $S^t$, where the negatives come from a dynamic diversity-enhanced negative text queue.
%
%
%
In the report-learning stage, the visual encoder, structural queries, and patch selection layer are frozen. 
The patch selection layer is used to select the most informative image patch embeddings $T^s$ for each primary structure, in addition to extracting $ S^v$.
Then, a text decoder, which receives $T^s$ and $S^v$ as input, is added and trained using a report-generation loss  $\mathcal{L}_\text{rg}$ for the target task of CTRG.
%


\vspace{3.5mm}\noindent\textbf{Structure-wise Visual Feature Extraction.}
Considering the large dimensions of 3D CT images and intricate details in CT reports, simply adopting 2D image-based semantic alignment strategies (e.g., subject-level contrastive learning \cite{liu2023improving}) to our scenario could be suboptimal. 
This is because local, subtle coherence between image and text could be ignored in global alignment, which is critical in fine-grained tasks like CTRG.
Huang et al. \cite{huang2021gloria} proposed an attention framework to learn local representations by contrasting image sub-regions and report words.
Medical CT reports are highly structured: they usually describe the same body parts from the same aspects---typically based on anatomical structures.
Therefore, we propose to parse the reports at the \textit{structure} level instead of word-wise, and align them with CT image patch embeddings accordingly.
Below, we describe the image processing part first and will deal with the reports shortly.
%
%

As shown in Fig. \ref{framework}(a), we extract the critical image patch embeddings corresponding to each structure with a set of structure-specific visual queries $Q^v=[\bm{q}^v_i]_1^{N^s}$, where $N^s$ is the number of considered structures.
Denoting the embeddings of all image patches by $F^v=[\bm{v}_j]_1^{N^v}$, where $N^v$ is the number of image patches, their cross-attention with the visual queries are computed:
\begin{equation}\label{eq:QK}\small
    A^v = \operatorname{softmax}[Q^vW^v_0(F^vW^v_1)^T],
\end{equation}
where $W^v_0$ and $W^v_1$ are linear projection matrices, and $A^v \in \mathbb{R}^{N^s\times N^v}$ is a pair-wise similarity matrix between the structure queries and image patches.
%
Then, we obtain a set of visual, subject-specific structure observation embeddings $S^v$:
\begin{equation}\small
    S^v = [\bm{s}^v_i]_1^{N^s} = A^v (F^v W^v_2),
\end{equation}
where $W^v_2$ is a linear projection matrix.
$S^v$ will be used for contrast shortly.

\vspace{3.5mm}\noindent\textbf{Structure-wise Textual Token Extraction.}
We employ a pretrained text encoder \cite{hamamci2024foundation} to extract embeddings for the textual structure observations.
This encoder processes sentences that describe each specific structure, converting them into embeddings. 
The resulting embeddings, specifically the outputs of the class tokens for each structure, are collected as textual subject-specific structure observation tokens, denoted by $S^t=[\bm{s}^t_i]_1^{N^s}$ (Fig. \ref{framework}(b)).
Since we aim to align the visual encoder with a well-trained text encoder, we freeze the text encoder's parameters during training.
To identify the sentences describing each structure, we implement a keyword-matching method: if a sentence contains any predefined keyword for a structure, it is considered to describe the structure.
We also define an ``others'' category for sentences with no predefined keyword.
Thus, not only does our report parsing rely on easily accessible high-level prior knowledge, but it is also straightforward to implement.

\vspace{3mm}\noindent\textbf{Cross-modal Alignment.}
After obtaining the visual and textual subject-specific structure observation tokens, we align them by contrastive learning.
Following \cite{li2021align}, we learn a similarity function $\operatorname{sim}(\bm{s}^v_i, \bm{s}^t_i)=g_v(\bm{s}^v_i)^Tg_t(\bm{s}^t_i)$, where $g_v$ and $g_t$ are linear transformations mapping the embeddings to normalized lower-dimensional ($512$-d) representations. 
Then, we calculate the softmax-normalized image-to-text similarity:
\begin{equation}\label{sim}\footnotesize
    \begin{aligned}
    {p}_m^{v2t}(\bm{s}_i^v) = \frac{\exp[\operatorname{sim}(\bm{s}^v_i, \bm{s}^t_i) / \tau]}
    {\sum_{m=1}^{N^sN^q} \exp[\operatorname{sim}(\bm{s}^v_i, \bm{s}^t_m)/\tau]} ,
    \end{aligned}
\end{equation}
where $\tau$ is a learnable temperature parameter, $\bm{s}^t_m$ is a sample in a queue $Q_a$ of textual tokens (Fig. \ref{framework}(b)), and $N^q$ is the queue length.
Then, our structure observation driven image-text contrastive loss is defined as:
\begin{equation}\label{loss_itc}\small
    \begin{aligned}
    \mathcal{L}_\text{so-itc}=\mathbb{E}_{(\bm{s}_i^v, \bm{s}_i^t)}
    \big[\operatorname{H}\big(\bm{y}^{v2t}(\bm{s}_i^v), \bm{p}^{v2t}(\bm{s}_i^v)\big)\big],
    \end{aligned}
\end{equation}
where $\operatorname{H}$ is the cross entropy, $\bm{p}=[p_m]_{m=1}^{N^sN^q}$, and $\bm{y}$ is the one-hot ground truth (1 for the positive pair and 0s for negative pairs).
We define the positive pair as the visual and textual observation embeddings of the same subject and structure.

\vspace{3mm}
\noindent\textbf{Soft Pseudo Targets.}
The one-hot label and training objective in Eqn. (\ref{loss_itc}) above penalize all negative pairs regardless of their correctness.
Negative text for an image may also match the image’s content, and vice versa. 
For example, the lung description of a patient with COVID-19 may also describe another patient's lung with COVID-19, although the text and image are from different subjects.
To mitigate this issue, we devise a soft pseudo target based on the similarity between textual observation tokens.
Concretely, the text-text similarity between two textual structure observation tokens is computed by $\operatorname{sim}'(\bm{s}^t_i, \bm{s}^t_m)=g_t(\bm{s}^t_i)^Tg_t(\bm{s}^t_m)$.
Then, the soft pseudo target $\bm{q}^{t2t}$ can be obtained by replacing the similarity function $\operatorname{sim}$ in Eqn. (\ref{sim}) with $\operatorname{sim}'$.
Finally, we impose a Kullback-Leibler (KL) divergence loss between the normalized image-to-text similarities $\bm{p}^{v2t}$ and the soft target $\bm{q}^{t2t}$:
\begin{equation}\label{loss_itc_mod}\small
    \mathcal{L}_\text{so-kl}=\mathbb{E}_{(\bm{s}_i^v, \bm{s}_i^t)}
    \big[\operatorname{KL}\big(\bm{q}^{t2t}(\bm{s}_i^t)|| \bm{p}^{v2t}(\bm{s}_i^v)\big)\big].
\end{equation}
The intuition is that, for a pair of visual and textual structure tokens of the same subject and structure, their similarities to other textual tokens are expected to be close, given the two modalities are well aligned.

The complete structure observation driven pretraining objective is: 
\begin{equation}\label{loss_so_pre}\small
    \mathcal{L}_\text{so-pre}=(1-\alpha)\mathcal{L}_\text{so-itc} + \alpha\mathcal{L}_\text{so-kl},
\end{equation}
where $\alpha$ is set to 0.5 empirically.
Compared with global alignment, our $\mathcal{L}_\text{so-pre}$ aligns the structure-wise semantics between the images and reports to learn fine-grained representations.

\vspace{3.5mm}\noindent\textbf{Diversity-Enhanced Queue.}
During structure learning, we maintain a queue of textual structure observation tokens.
To improve the efficiency of contrastive learning, we propose storing the most informative samples in the queue. 
Specifically, for each token $\bm{s}^t_i$ in a batch, we calculate the sum of its text-text similarities to all tokens in the queue: $S_i=\operatorname{sum}(g_t(\bm{s}^t_i)^Tg_t(\bm{s}^t_m))$.
If its $S_i$ value is smaller than the largest $S$ of all tokens in the queue, it is enqueued, and the queue token with the largest $S$ is dequeued.
For practical implementation, we store the $S$ values for tokens already in the queue instead of recomputing them.
This selective process ensures that the queue contains diverse and informative samples for discriminative and contrastive learning while keeping the queue size relatively small to reduce computational overhead.
\vspace{3.5mm}\noindent\textbf{Report Learning and Inference.}
The preceding stage serves as a high-level medical prior informed, image-text pretraining in which a fine-grained, local structural image representation is learned.
In the second stage (Fig. \ref{framework}(c)), we freeze the image encoder, visual queries, and patch-selection layers and add and train a text decoder for report generation (note the text encoder in the preceding stage is no longer used).
Our framework is agnostic to the exact text decoder used and applies to various text generation models.
In this work, we experiment with a BERT decoder \cite{li2022blip} and a large language model (LLM) LLaMA2-7B \cite{touvron2023llama}.\footnote{For LLaMA2-7B, we use LoRA \cite{hu2021lora} for parameter-efficient fine-tuning and a two-layer multilayer perception for visual feature projection.
}
Formally, the model is trained with the chained next-token prediction objective:
$\mathcal{L}_\text{rg} = -{\sum}_{k=1}^{N^t} \log P(\bm{t}_k|\bm{t}_{1:k-1})$,
where $P$ is the probability of predicting the next token $\bm{t}_k$ conditioned on preceding ones of the target report.

To provide detailed information for comprehensive report statement generation, we consider two types of tokens as joint input to the text decoder: first, the image structural representations $S^v$ to convey general information about each structure; second, the selected most informative image patch embeddings $T^s$ to provide detailed information for each structure.
Specifically, based on $A^v$, {\color{blue}the pair-wise similarity matrix between the structure queries and image patches computed in Eqn. (\ref{eq:QK})}, we identify and preserve $K$ image patch embeddings most similar to each structural query, resulting in a total of $K \times N^s$ patch embeddings selected, denoted by $T^s$. $K$ is set to 10 empirically (cf. Tables \ref{tabs:clusNum} and \ref{ablas:ce}).

\section{Experiments}

\subsubsection{Datasets.}
We conduct experiments on two public CT report generation datasets.
\textbf{1) CT-RATE} \cite{hamamci2024foundation} includes 25,692 non-contrast chest CT volumes with corresponding reports of 21,304 unique patients.
Following \cite{hamamci2024foundation}, we standardize all CT volumes to a uniform voxel spacing of $0.75\times0.75\times1.5$ mm\textsuperscript{3}. 
The volumes are either center-cropped or padded to a consistent resolution of $480\times480
\times240$ voxels.
We use the official training set (24,128 volumes/20,000 patients) for training, and further split the official test set into a validation (360 volumes/300 patients) and a testing set (1,204 volumes/1,004 patients).
The validation set is only used for model optimization, e.g., in ablation studies, whereas the test set is used for final performance reporting and comparison with other methods.
%
%
\textbf{2) CTRG-Chest-548K} \cite{tang2024work} includes 1,804 chest CT volumes and corresponding reports.
We use the same data split as in \cite{chen2024dia}: 60\% for training, 20\% for validation, and 20\% for testing.
Following \cite{chen2024dia}, each volume is resized to $256\times256$ voxels.
%
%
For each case, we use the CT image and Findings section of the English report.

\vspace{3.5mm}\noindent\textbf{Evaluation Metrics.}
{Following common practice in the literature (e.g., \cite{chen2024dia,jin2024promptmrg}), we employ both natural language generation (NLG) and clinical efficacy (CE) metrics for evaluation. 
NLG metrics include BLEU \cite{papineni2002bleu}, METEOR \cite{banerjee2005meteor}, and ROUGE-L \cite{lin2004rouge}. 
CE metrics include precision, recall, and F1 score following \cite{jin2024promptmrg} and \cite{nicolson2023improving}. 
For CT-RATE, we use the officially provided text classifier to extract 18 distinct types of abnormalities to calculate CE metrics.
For CTRG-Chest-548K, following \cite{chen2024dia}, we utilize a pretrained report labeler, CheXbert \cite{smit2020chexbert}, to extract labels, which remains effective in labeling CT reports as shown by \cite{chen2024dia}. 
It should be noted that while some of the compared methods achieved high NLG metrics in our experiments, the sentences generated are highly repetitive and lack diversity and diagnostic value.
This may result from the highly structured format of the reports in the two datasets.
Thus, the generated reports closely follow the ground truth regarding wording, description sequence, etc., but fail to capture the diagnostic information expected for a medical report. 
Therefore, we mainly discuss the models' performance based on CE metrics below.
%

\vspace{3.5mm}\noindent\textbf{Implementation.}
The proposed framework is implemented using PyTorch (2.0.0) \cite{paszke2019pytorch} and trained with four NVIDIA Tesla V100 GPUs with 32GB memory each. 
The optimizer is AdamW \cite{loshchilov2018decoupled}, with learning rates of $10^{-5}$ and $10^{-4}$ for structure and report learning. 
We set the warm-up ratio to 10$\%$ and use the linear learning rate scheduler after warm-up. 
The model is trained for 100K and 30K steps for structure and report learning, with a batch size of 8.
The number of visual structural queries $N^s$ is set to 10, corresponding to lung, trachea and bronchie, mediastinum and heart, esophagus, pleura, bone, thyroid, breast, abdomen, and others (e.g., thoracic cavity, prostate and so on, which are only mentioned in few reports), following the hierarchy of anatomical regions in \cite{zhang2024radgenome}.
%
The length $N^q$ of the diversity-enhanced queue for the structure-wise image-text contrasting is empirically set to 1000 for each structure.
Note that we only need to parse the original reports into structure-corresponding sentences for the structure-learning stage.
The original reports are used for report learning and performance evaluation.
The same preprocessing procedures are implemented for all compared methods for a fair comparison.
{\color{blue}Our implementation, including the alignment data, is available at: https://github.com/ccarliu/CTRG.}

\vspace{3.5mm}\noindent\textbf{Comparison with State-of-the-Art (SOTA) Methods.}
We compare our framework to various SOTA methods, including those proposed for chest X-ray: R2Gen \cite{chen2020generating}, R2GenCMN \cite{chen2022cross}, M2KT \cite{yang2023radiology}, R2GenGPT \cite{wang2023r2gengpt}, PromptMRG \cite{jin2024promptmrg}; 
radiology pretraining methods: RadFM \cite{wu2023towards}, CT-CLIP \cite{hamamci2024foundation}, GLoRIA \cite{huang2021gloria};
and specialized CT report generation methods: CT2Rep \cite{hamamci2024ct2rep}, SL-DG \cite{tang2024work}, Dia-LLaMA \cite{chen2024dia}. 
Unless otherwise specified, we use the codes released by the authors. 
{\color{blue}For the pretraining methods, we freeze the pretrained image encoders (i.e., checkpoints released by the authors \cite{wu2023towards,hamamci2024foundation} or trained with author-released codes \cite{huang2021gloria}), and add and train the same BERT decoder \cite{li2022blip} as our framework for report generation.}
For methods originally designed for 2D images, we replace their image encoders with the same one as ours to extract CT image representations.
We optimize the performance of all compared methods on the validation data.

\begin{table}[!t]
  \centering
  \caption{The performance of our model compared with SOTA methods on the test sets of CT-RATE \cite{hamamci2024foundation} (left) and CTRG-Chest-548K \cite{tang2024work} (right). 
  *: cited from the original paper. 
  $\dag$: our model with CT representation learned on CT-RATE.
}\label{tab:comparison1}
  \begin{minipage}{0.49\textwidth}
    \centering
    \vspace{1.5mm}
    \setlength{\tabcolsep}{0.8mm}
    \begin{adjustbox}{width=\textwidth}
    \begin{tabular}{c|ccc|cccc}
    \hline
    \multicolumn{8}{c}{\textit{CT-RATE} \cite{hamamci2024foundation}}\\
    \hline
     & \multicolumn{3}{c|}{\textbf{CE Metrics}} & \multicolumn{4}{c}{\textbf{NLG Metrics}}  \\
    \textbf{Method} &  \textbf{Pre.} & \textbf{Rec.} & \textbf{F1} & \textbf{BL-1} & \textbf{BL-4} & \textbf{MTR} & \textbf{RG-L} \\
    \hline
    R2Gen \cite{chen2020generating}  & 0.140 & 0.026& 0.043 &  0.426 & 0.241 & 0.248 & 0.333 \\
    R2GenCMN \cite{chen2022cross} & 0.187 & 0.094& 0.091  & 0.431 & 0.243 & 0.266 & 0.350 \\
    M2KT \cite{yang2023radiology} & {0.323} & 0.120 & 0.156 & 0.456 & \textbf{0.261} & {0.255} & 0.348 \\
    R2GenGPT \cite{wang2023r2gengpt} &{0.326}&0.205&0.228&0.166&0.091&0.116&0.161 \\  
    PromptMRG \cite{jin2024promptmrg}  & \underline{0.353} & {0.286} & {0.288} & 0.434 & 0.186 & 0.207 & 0.361 \\
    \hdashline
    RadFM \cite{wu2023towards}   & 0.143 & 0.154 & 0.166 & 0.456 & \underline{0.259} & \textbf{0.278} & \underline{0.382} \\ 
    CT-CLIP \cite{hamamci2024foundation} & 0.217 & 0.129& 0.145& 0.455& \underline{0.259} & \underline{0.277} & \textbf{0.383} \\ 
    GLoRIA \cite{huang2021gloria} &0.298&{0.251}&0.240&0.448&0.220&0.270&0.346 \\ 
    \hdashline
    CT2Rep \cite{hamamci2024ct2rep}  & 0.206 & 0.073 & 0.099 & {0.447} & {0.251} & 
    {0.275} & {0.377} \\ 
    \hline
    Ours-BERT  & 0.321  & \textbf{0.354} & \textbf{0.310} & \textbf{0.486} & 0.254 & {0.275} & 0.351 \\ 
    Ours-LLaMA & \textbf{0.354} & \underline{0.328} & \underline{0.308} &\underline{0.481}&0.225&0.266  &0.318  \\ 
    \hline
    \end{tabular}
    \end{adjustbox}
    
  \end{minipage}
  \hspace{0.003\textwidth} 
  \begin{minipage}{0.49\textwidth}
    \centering
    \setlength{\tabcolsep}{0.8mm}
    \begin{adjustbox}{width=\textwidth}
    \begin{tabular}{c|ccc|cccc}
    \hline
    \multicolumn{8}{c}{\textit{CTRG-Chest-548K} \cite{tang2024work}}\\
    \hline
     & \multicolumn{3}{c|}{\textbf{CE Metrics}} & \multicolumn{4}{c}{\textbf{NLG Metrics}}  \\
    \textbf{Method} &  \textbf{Pre.} & \textbf{Rec.} & \textbf{F1} & \textbf{BL-1} & \textbf{BL-4} & \textbf{MTR} & \textbf{RG-L} \\
    \hline
    R2Gen \cite{chen2020generating}  & 0.211 & 0.124 & 0.146 & 0.332 & 0.229 & 0.217 & \textbf{0.473} \\
    R2GenCMN \cite{chen2022cross}   & 0.152 & 0.108 & 0.115 & 0.368 & 0.244 & 0.224 & \underline{0.451} \\
    M2KT \cite{yang2023radiology} & 0.227 & 0.114 & 0.146 & 0.453 & 0.229 & 0.262 & 0.384 \\
    R2GenGPT \cite{wang2023r2gengpt} &0.273&0.130&0.175&0.259&0.147&0.263&0.275 \\ 
    PromptMRG \cite{jin2024promptmrg}  & 0.311 & 0.347 & 0.301 & 0.467 & 0.236 & 0.231 & 0.383 \\
    \hdashline
    RadFM \cite{wu2023towards}   & 0.395 & 0.353 & 0.341 & 0.471 & 0.253 & 0.244 & 0.396 \\ 
    CT-CLIP \cite{hamamci2024foundation} &0.318&0.228&0.246&0.421&0.267&0.263&0.443 \\ 
    GLoRIA \cite{huang2021gloria} &{0.411}&0.376&{0.358}&0.453&0.282&\textbf{0.280}&0.439 \\ 
    \hdashline
    CT2Rep \cite{hamamci2024ct2rep}  & 0.223 & 0.135 & 0.125 & 0.432 & 0.251 & 0.259 & 0.408 \\ 
    SL-DG* \cite{tang2024work}  & - & - & - & - & 0.237 & 0.219 & 0.438 \\
    Dia-LLaMA* \cite{chen2024dia} & {0.421} & {0.387} & {0.372} & \textbf{0.512} & \textbf{0.296} &{0.263} & {0.422} \\ 
    \hline
    Ours-BERT  & \underline{0.434} & \textbf{0.413} & \textbf{0.393} & \underline{0.501} & \underline{0.294} & \underline{0.265} & {0.437}  \\ 
    Ours-LLaMA  &  \textbf{0.435} & \underline{0.410} & \underline{0.387} & 0.340 & 0.183 & \textbf{0.280} & 0.317 \\ 
    \hdashline
    Ours-BERT\textsuperscript{\dag} & {0.503} & {0.496} &{0.468} & 0.511 & 0.297 & 0.266 & 0.419   \\  
    \hline
    \end{tabular}
    \end{adjustbox}
    
  \end{minipage}
\end{table} 

Table \ref{tab:comparison1} shows the performance of all compared methods.
As we can see, on both datasets, R2Gen, R2GenCMN, and M2KT yield generally low CE metrics, despite their reasonable NLG metrics.
R2GenGPT shows some improvements in CE metrics by leveraging the capabilities of a large language model (LMM), but it has notably low NLG metrics.
We conjecture this is because their global alignment strategies neglect vital sub-regions of CT images and thus fail to capture fine-grained features.
PromptMRG achieves significant improvement in CE metrics after incorporating diagnostic classification results.
Among the three pretrained models (RadFM, CT-CLIP, and GLoRIA), GLoRIA performs the best. 
It likely benefits from its localized correspondence modeling between images and reports, which can capture more detailed information for report generation.
Among the models with specific designs for CT report generation (CT2Rep, 
\begin{figure*}[t]
    \centering
    \includegraphics[trim=0 0 0 0, clip, width=.95\linewidth]{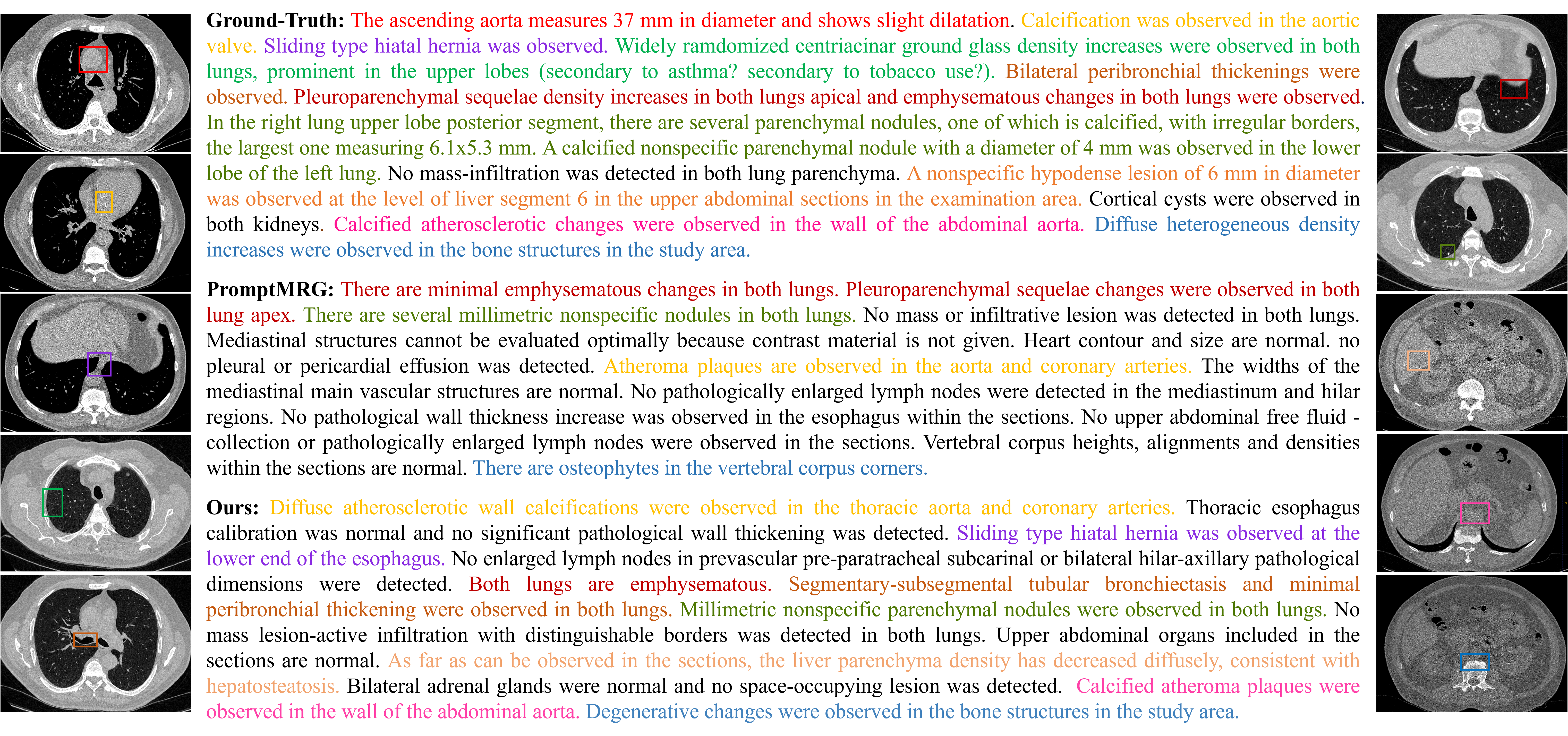}\vspace{-2mm}
    \caption{Ground truth and example reports generated by {PromptMRG} \cite{jin2024promptmrg} and our method for a case in CT-RATE \cite{hamamci2024foundation}.
    Colored texts and color- and order-matching boxes in CT slices indicate clinic-relevant findings in the ground truth.
    Our method not only produces sentences with fine details but also identifies more observations than PromptMRG.
    CT slices are shown with varying windows to highlight lesions.}
    \label{fig:report_vis}
\end{figure*}
SL-DG, and Dia-LLaMA), Dia-LLaMA yields decent performance. Its incorporation of diagnostic information and the capability of the LLM may benefit it.
%
%
Finally, both implementations of our framework, using BERT (Ours-BERT) or LLaMA2-7B (Ours-LLaMA) as the text decoder, achieve superior recall and F1 scores to other methods across the two datasets.
Meanwhile, the precision scores of Ours-LLaMA are the highest, too.
Interestingly, while Ours-BERT is also competent in the NLG metrics on both datasets, Ours-LLaMA is generally less effective in NLG metrics, especially on CTRG-Chest-548K.
We conjecture that this discrepancy is due to the limited data available for effectively fine-tuning the LLM (CTRG-Chest-548K has only 1262 cases for training).
It is worth emphasizing that our framework is built upon the easily accessible high-level prior knowledge of what structures a chest CT depicts. In contrast, PromptMRG, SL-DG, and Dia-LLaMA require term-level manual annotations for training.
These results demonstrate the strong capability of our method in CTRG.

Figure \ref{fig:report_vis} shows examples of generated CT reports for a case in the CT-RATE dataset.
Compared with PromptMRG, our method generates more comprehensive diagnostic reports for various thoracoabdominal organs. 
In contrast, PromptMRG is more oriented to the abnormality annotations it uses for training and primarily includes abnormalities related to the heart and lungs (most of the abnormality annotations in this dataset are focused on the heart and lungs).
%

\vspace{3.5mm}\noindent\textbf{Transfer CT Representation.}
To assess the quality and generalizability of the CT representation learned by our structure-learning framework, we transfer the representation learned on CT-RATE to CTRG-Chest-548K for subsequent report learning.
The bottom row of Table \ref{tab:comparison1} demonstrates substantial performance improvements upon the representation learned on the small CTRG-Chest-548K dataset in all CE metrics and comparable performance for all NLG ones.
We attribute the improvements to the better quality of the representation learned on the larger CT-RATE dataset.
These results also validate the efficacy of our methodology in learning CT representation that is generalizable across domains.

\begin{table*}[t]
\centering
\caption{Ablation study on the {\textit{validation}} set of CT-RATE \cite{hamamci2024foundation}.
}\label{tab:ablation}
\begin{adjustbox}{width=.90\textwidth}
\begin{tabular}{c|ccc|cc|ccc|cccc}
\hline
Ablation & \multicolumn{3}{c|}{pretrain} &\multicolumn{2}{c|}{Fine-tune} & \multicolumn{3}{c|}{CE Metrics $\uparrow$}  & \multicolumn{4}{c}{NLG Metrics $\uparrow$} \\ \cline{2-13}
config. & $\mathcal{L}_\text{so-itc}$ & $\mathcal{L}_\text{so-kl}$ & Diversity Queue & $S^v$ & $T^s$ & Pre. & Rec.& F1&BL-1 & BL-4          & MTR            & RG-L                                 \\ \hline
(a) & $\times$                    & $\times$                    &   $\times$ & $\checkmark$ & $\times$  & 0.221 & 0.115 & 0.124 & \textbf{0.511} & 0.\textbf{286} & 0.272 & \textbf{0.374}   \\ 
(b)      & \checkmark                  & $\times$       & $\times$      & $\checkmark$ & $\times$       &  0.307 & 0.317 & 0.300    & 0.497 & 0.266 &0.272 & 0.357   \\ 
(c)      & \checkmark                     & \checkmark           & $\times$& $\checkmark$& $\times$       & \underline{0.325} & 0.318 & 0.306 & 0.502 & {0.273} & \underline{0.273} & 0.364   \\ 
(d)      & \checkmark                     & \checkmark           & \checkmark & $\checkmark$& $\times$       & \textbf{0.334} & \underline{0.323} & \underline{0.309} & 0.488 & 0.257 & 0.268 & 0.348    \\\hline 
Full     & \checkmark                   &  \checkmark           & \checkmark & \checkmark & \checkmark                  & \textbf{0.334} & \textbf{0.356} & \textbf{0.320}  & \underline{0.509} & \underline{0.282} & \textbf{0.277} & \underline{0.371} \\\hline     
\end{tabular}
\end{adjustbox}

\end{table*}



\vspace{3.5mm}\noindent\textbf{Ablation Study.}
We conduct ablative experiments to validate the efficacy of our framework's novel designs on the {\textit{validation}} set of CT-RATE \cite{hamamci2024foundation} with the BERT text decoder \cite{li2022blip} (Table \ref{tab:ablation}).
The baseline (a) is our backbone network trained with $\mathcal{L}_\text{rg}$ alone, yielding the worst CE metrics.
Row (b) adds the proposed structure learning with the structure observation-driven image-text contrastive loss $\mathcal{L}_\text{so-itc}$.
It brings noticeable improvements upon the baseline, with at least 8.6\% improvements for all CE metrics.
Row (c) further enhances performance by introducing text-text similarity-based soft pseudo targets ($\mathcal{L}_\text{so-kl}$), improving CE metrics by 1.8\%, 0.1\%, and 0.6\%.
The diversity-enhanced negative queue also boosts performance in row (d), with improvements of 0.9\%, 0.5\%, and 0.3\% in CE metrics. 
These two ablations validate the effectiveness of our proposed text-text similarity-based soft pseudo targets for contrastive learning, and the diversity-enhanced negative queue in learning diagnostic-related structure observation embeddings for report generation.
%
Eventually, our full model achieves the best performance for all CE metrics with $\mathcal{L}_\text{so-itc}$, $\mathcal{L}_\text{so-kl}$ and the diversity queue for structure learning, and $S^v$ and $T^s$ for report learning.
%
%

\begin{table}[!t]
  \centering

  \begin{minipage}{0.505\textwidth}
    \centering
    \setlength{\tabcolsep}{0.8mm}
    \caption{Experimental results on the \textit{validation} set of CT-RATE \cite{hamamci2024foundation} in CE metrics, with the BERT text decoder \cite{li2022blip}.
    Left: varying $\alpha$ (weight of $\mathcal{L}_\text{so-kl}$) with $K=10$. 
    Right: varying the number $K$ of selected image patch embeddings for each structure with $\alpha=0.2$.}%
    \label{tabs:clusNum}
    \begin{adjustbox}{width=\textwidth}
    \begin{tabular}{c|ccccc|ccc}
    \cline{1-4}\cline{6-9}
    $\alpha$ & Pre.          & Rec.            & F1   &    & $K$     &    Pre.          & Rec.            & F1                       \\ \cline{1-4}\cline{6-9}
    0.0                 &  0.318&0.335&0.307&&0&\textbf{0.338} & {0.323} & {0.309} \\
    0.1                 &  \textbf{0.342}&0.331&0.313&&5& {0.331} & {0.332} & {0.310} \\
    0.2  &   {0.334} & \textbf{0.356} & \textbf{0.320}    &&10        & {0.334} & \textbf{0.356} & \textbf{0.320} \\
    0.3                 &  0.321&0.334&0.306&&15& 0.328 & 0.342 & 0.307 \\
    0.4                 &  0.312&0.328&0.301&&20& 0.322&0.318&0.303\\ \cline{1-4}\cline{6-9}
    \end{tabular}
    \end{adjustbox}
     \end{minipage}
     \hspace{0.01\textwidth}
     \begin{minipage}{0.46\textwidth}
        \centering
    \caption{
    CE metrics (on the \textit{validation} set), training memory consumption, and GMACs comparison on the CT-RATE \cite{hamamci2024foundation} dataset by varying the number $K$ of selected image patch embeddings for each structure.
    LLaMA2-7B \cite{touvron2023llama} is used as the text decoder.}\vspace{-1mm}
    \setlength{\tabcolsep}{.8mm}
    \begin{adjustbox}{width=.83\width}
    \begin{tabular}{c|ccc|cc}
    \hline
    $K$ & Pre.          & Rec.            & F1     &  Mem. (G)  &  GMACs          \\ \hline
    0                 &  0.341&0.298&0.291 &23.8&8730.9\\
    5                 &  {0.353}&{0.316}&{0.306} &24.5&9733.8 \\
    10  &   \textbf{0.355}&\textbf{0.322}&\textbf{0.309} &25.1&10760.8  \\
    15                 &  0.339&0.321&0.302 &25.8&11781.6\\
    20                 &  0.318&0.311&0.289 &26.5& 12806.5\\ \hline
    \end{tabular}
    \end{adjustbox}

\label{ablas:ce}
 \end{minipage}
 \end{table}

\vspace{3.5mm}\noindent\textbf{Validate Designs of Our Method.}
Based on the validation set of CT-RATE, we determine 1) the optimal weight $\alpha$ for $\mathcal{L}_\text{so-pre}$ (Eqn. (6)) and 2) the optimal number $K$ of selected image patch embeddings $T^s$ for each structure, with the BERT text decoder \cite{li2022blip}.
We fix either while varying the other to reduce the search space.
The results are shown in Table \ref{tabs:clusNum}. 
Although the results look fairly stable, we select $\alpha=0.2$ and $K=10$ for final performance evaluation and comparison with other methods on the test data, due to their overall high performance in CE metrics compared with alternative values.

\vspace{3.5mm}\noindent\textbf{Effects of Patch Selection.}
Table \ref{ablas:ce} presents the CE metrics, training memory consumption, and GMACs (giga multiply-accumulate operations per second) for our model using LLaMA2-7B \cite{touvron2023llama} as text decoder. 
Our model demonstrates commendable performance with $K=10$, i.e., 110 total visual tokens (comprising 100 selected tokens $T^s$ and 10 structure observation embeddings $S^v$) are fed to the decoder.
This is significantly fewer than the original 4096 visual tokens produced by the visual encoder, which is beyond the limit of our hardware (maximum $\sim$600 visual tokens when fine-tuning LLaMA2-7B with LoRA).
Therefore, our structure-observation-driven visual feature extraction not only makes contrastive learning more focused on the main structures in chest CTs, but also effectively reduces computational overhead.

\begin{wraptable}{r}{5cm}
\centering\vspace{-6mm}
\caption{Report-to-volume retrieval performance on the test set of CT-RATE \cite{hamamci2024foundation}.
*: cited from \cite{hamamci2024foundation}.
}\label{retrieval}\vspace{1mm}
\begin{adjustbox}{width=.72\width}
\begin{tabular}{c|ccc}
\hline
Method & Rec.@10          & Rec.@50   & Rec.@100   \\ \hline
CT-CLIP \cite{hamamci2024foundation}*             &  0.040&0.144&0.235\\
Ours w/o $\mathcal{L}_\text{so-kl}$               &  {0.047}&{0.175}&{0.279} \\
Ours  &   \textbf{0.049} & \textbf{0.189} & \textbf{0.296}\\
 \hline
\end{tabular}
\end{adjustbox}\vspace{-5mm}
\end{wraptable}
\vspace{3.5mm}\noindent\textbf{Report to Volume Retrieval.}
Following \cite{hamamci2024foundation}, we evaluate report-to-volume retrieval performance by computing the cosine similarity between the embeddings of target volumes and query texts.
We report Recall@10, 50, and 100 and compare with CT-CLIP \cite{hamamci2024foundation} in Table \ref{retrieval}.
Both variants of our method demonstrate superior retrieval performance compared with CT-CLIP, indicating that our structure observation-driven image-text contrastive learning successfully captures the fine-grained subtle coherence between CT volumes and reports.
Additionally, the declined performance of our model without 
$\mathcal{L}_\text{so-kl}$ (Ours w/o $\mathcal{L}_\text{so-kl}$) confirms the effectiveness of $\mathcal{L}_\text{so-kl}$ in enhancing cross-modal representation by mitigating the false-negative issue.

\section{Conclusion}

This paper presented a novel framework for learning to generate medical reports for 3D CT images.
The framework featured effective structure-wise image-text learning built on high-level medical prior knowledge.
Extensive experiments on chest CT demonstrated its superior performance to existing SOTA approaches and the efficacy of its novel designs.
We plan to extend our framework for other volumetric imaging data.
A limitation of this work was that the NLG metrics of our framework with the LLaMA2-7B decoder were not as good as those with the BERT decoder.
This could result from the insufficiency of conventional NLG metrics in evaluating the more capable LLMs nowadays.
We expect to assess LLMs' performance with better tools in the future.

\begin{credits}
\subsubsection{\ackname} This work was supported by the National Natural Science Foundation of China (Grant No. 62371409) and Fujian Provincial Natural Science Foundation of China (Grant No. 2023J01005).
\end{credits}

\bibliographystyle{splncs04}
\bibliography{references}
%
%
%
%

\end{document}